\documentclass[lettersize,journal]{IEEEtran}
\usepackage{amsmath,amsfonts}
\usepackage{algorithmic}
\usepackage{algorithm}
\usepackage{array}
\usepackage[caption=false,font=normalsize,labelfont=sf,textfont=sf]{subfig}
\usepackage{textcomp}
\usepackage{stfloats}
\usepackage{url}
\usepackage{verbatim}
\usepackage{graphicx}
\usepackage{cite}
\usepackage{multirow}
\usepackage{booktabs}
\usepackage{xcolor}
\usepackage{url}
\definecolor{revisionblue}{RGB}{0,0,200}

\begin{document}

\title{Full-Body Golf Swing Kinematic Reconstruction \\From a Smartwatch IMU}

\author{Yuanshuo~Tan, Kezhe~Zhu, Xiujie~Sun, Chunping~Liang, Shuoyang~Zhu, Chenquan~Xu, Xinda~Jia, Licheng~Zhong, Huiming~Pan, Yinri~Jin, Chang~Liu, Bo~Xiao, Shenglong~Le, Bryndan~W.~Lindsey, and Peter~B.~Shull
\thanks{This work was supported in part by the National Natural Science Foundation of China under Grant W2441018 and in part by the Shanghai Sports Science and Technology Project under Grant 26Q009. (Corresponding authors: Bryndan~W.~Lindsey; Shenglong~Le.)}
\thanks{Yuanshuo~Tan, Kezhe~Zhu, Xiujie~Sun, Chunping~Liang, Chenquan~Xu, Xinda~Jia, Huiming~Pan, Bryndan~W.~Lindsey, and Peter~B.~Shull are with the State Key Laboratory of Mechanical System and Vibration, School of Mechanical Engineering, Shanghai Jiao Tong University, Shanghai 200240, China (e-mail: tanyuanshuo@sjtu.edu.cn; kezhe\_zhu@sjtu.edu.cn; sxj2003@sjtu.edu.cn; pp2003@sjtu.edu.cn; xiaoqin@sjtu.edu.cn; xinda\_jia@sjtu.edu.cn; panhuiming@sjtu.edu.cn; blindse3@sjtu.edu.cn; pshull@sjtu.edu.cn).}
\thanks{Shuoyang Zhu is with the Key Laboratory for Power Machinery and Engineering of the Ministry of Education, Shanghai Jiao Tong University, Shanghai 200240, China (e-mail: zsy\_0308@sjtu.edu.cn).}
\thanks{Licheng Zhong is with the Department of Computer Science, National University of Singapore, Computing 1, 13 Computing Drive, Singapore (e-mail: zlicheng@u.nus.edu).}
\thanks{Yinri Jin is with the School of Physical Education, Shanghai University of Sport, Shanghai 200438, China (e-mail: jinri@sus.edu.cn).}
\thanks{Chang Liu, Bo Xiao, and Shenglong Le are with the Department of Physical Education, Shanghai Jiao Tong University, Shanghai 200240, China (e-mail: liuc@sjtu.edu.cn; xiaobo33@sjtu.edu.cn; longsonlok@sjtu.edu.cn).}
\thanks{Shenglong Le is also with the Department of Physical Therapy, Taihe Hospital, Hubei University of Medicine, Shiyan 442099, China, and also with the Center for Diabetes Rehabilitation Research, Taihe Hospital, Hubei University of Medicine, Shiyan 442099, China.}}

\markboth{IEEE TRANSACTIONS ON INSTRUMENTATION AND MEASUREMENT}%
{Tan \MakeLowercase{\textit{et al.}}: Full-Body Golf Swing Kinematic Reconstruction From a Smartwatch IMU}


\maketitle

\begin{abstract}
Quantitative measurement of the golf swing is critical for evaluating technique and enabling individualized feedback. However, existing methods are impractical to use on the golf course: optical motion capture is laboratory-bound, camera-based methods require impractical camera placement, and multi-sensor inertial measurement unit (IMU) systems require multi-segment setup and calibration. We thus propose a single wrist-worn IMU approach for estimating full-body joint angles during golf swings. The proposed Wrist-IMU Temporal Kinematic Network (WIT-KinNet) combines IMU embedding, feature-wise linear modulation (FiLM), and temporal convolutional encoding. Thirty-six golfers, comprising beginner and skilled players, performed full, half, and quarter swings using seven club types: driver, 3-wood, 5-hybrid, 5-iron, 7-iron, 9-iron, and sand wedge. The proposed WIT-KinNet was evaluated under subject-wise cross-validation using synchronized smartwatch IMU data and ground-truth kinematics derived from an optical motion capture (OMC) system, with OMC-free IMU calibration and IMU-based swing segmentation. The proposed approach achieved a mean absolute error of $8.53\pm2.05^\circ$ across full-body joint angles. High temporal correlation was observed for pelvic rotation and upper torso rotation ($r=0.99$ for both), with X-factor and S-factor also showing strong correlations ($r=0.97$ for both). Linear mixed-effects models of the error revealed that swing amplitude significantly affected estimation error across all five trunk--pelvis variables ($p<0.05$). The results demonstrate the feasibility of full-body golf swing kinematic estimation from a commercial smartwatch, providing a potential solution for post-swing biomechanical analysis.

\end{abstract}

\section{Introduction}
\label{sec:Introduction}

\IEEEPARstart{G}{olf} is a globally popular sport, with approximately 66 million participants worldwide~\cite{o2024nutrition}. The golf swing is a highly technical, rapid, full-body movement requiring precise coordination of the pelvis, trunk, shoulders, arms, wrists, and lower limbs~\cite{hume2005role, nesbit2005three}. Kinematic variables including pelvis and trunk rotation, torso-pelvic separation angle (X-factor), shoulder and pelvic obliquity, and proximal-to-distal sequencing have been widely recognized as important indicators of golf swing proficiency~\cite{bourgain2022golf, chu2010relationship, callaway2012analysis, horan2012control, nesbit2005three, myers2008role, zheng2008kinematic, madrid2020associationt, cheetham2001importance, joyce2017most, meister2011rotational}. These variables directly reflect the magnitude and timing of body segment rotations, inter-segment angular relationships, and the sequencing of momentum transfer along the kinetic chain, and have consistently been shown to differ between golfers of varying skill levels and to correlate with key performance outcomes such as clubhead and ball speed~\cite{bourgain2022golf, cheetham2001importance, zheng2008kinematic, callaway2012analysis, mchugh2024kinematic, myers2008role}. 
Accordingly, measurement of full-body golf swing kinematics allows for objective quantification of performance-related variables that can be used by coaches and golfers to evaluate technique, identify inefficient movement patterns, and support performance training~\cite{zhou2022swing, smith2015golf, hume2005role}.

Three broad categories of methods have been investigated for golf swing kinematics analysis: laboratory-grade optical systems, camera-based markerless approaches, and body-worn inertial sensors. Optical motion capture (OMC) systems remain the gold standard for measuring three-dimensional joint angles across major body segments~\cite{cappozzo1995position, bourgain2022golf, kim2023validation}. Vision-based methods reduce the hardware burden by estimating two-dimensional or three-dimensional body keypoints directly from standard video footage, and have been adapted for golf swing event detection, comparison between learner and expert motion, and player and club trajectory reconstruction~\cite{mcnally2019golfdb, liao2022ai, ju2023golfmate, lee2024golfpose}.
In parallel, wearable inertial measurement unit (IMU)-based systems improve portability by attaching sensors to the body, reducing dependence on fixed camera infrastructure and controlled lighting~\cite{kim2020golf, kim2023validation, kim2024enhancing, cheon2020analysis, zhang2017sensor}. Multi-IMU systems with sensors placed on key body segments have been used to estimate three-dimensional joint angles at the wrist, shoulder, hip, and knee during golf swings~\cite{vo2025multisensor, zhu2023imu, li20243d}. Single-IMU approaches further reduce sensing requirements, with wrist-worn sensors used for swing phase segmentation and wrist trajectory reconstruction~\cite{kim2020golf, kim2024enhancing}. Collectively, these studies reflect a trend from laboratory-bound motion capture and camera-based systems toward wearable configurations with progressively reduced sensing requirements.

Despite these methodological advances, no existing approach is practical for routine, on-course swing analysis. OMC demands controlled laboratory conditions, multiple calibrated cameras, and trained operators, making field deployment infeasible~\cite{cappozzo1995position, bourgain2022golf}. Vision-based methods remain sensitive to viewpoint, lighting, body occlusion, and privacy constraints, generally restricting their use to indoor or driving-range settings~\cite{mcnally2019golfdb, liao2022ai, ju2023golfmate, lee2024golfpose}. Wearable IMU systems reduce the environmental dependencies, but most validated solutions rely on multiple sensors distributed across body segments, imposing non-trivial placement and calibration burdens that limit adoption during actual play~\cite{vo2025multisensor, kim2023validation, zhu2023imu, li20243d}. Single-IMU approaches have been explored to reduce sensing complexity, yet existing work has targeted coarse temporal segmentation or swing trajectory rather than full-body joint-angle reconstruction~\cite{kim2020golf, kim2024enhancing, cheon2020analysis}. More recently, full-body joint-angle estimation during the golf swing has been explored using a virtual wrist-worn inertial sensor~\cite{lauer2025learning}. Although the reported model achieved full-body joint angle errors averaging 4.0 $\pm$ 2.1$^\circ$ and joint position errors of 5.3 $\pm$ 1.1 cm, the wrist inertial signals and reference kinematics in that study were generated from a video-based reconstruction pipeline~\cite{shin2024wham}, rather than collected from a real smartwatch IMU and validated against marker-based OMC. The reliance on synthetic inertial inputs and video-derived reference kinematics limits the ecological validity of the reported findings to real-world smartwatch sensing and practical on-course deployment, as synthetic signals lack the noise, drift, and calibration artifacts inherent to real IMU hardware, potentially inflating reported accuracy. Consequently, a practically deployable solution for full-body golf swing kinematics estimation that is trained on and validated against real-world smartwatch IMU data and marker-based OMC remains to be developed. Such a solution should require only a single commercially available wrist-worn device, impose no environmental constraints, and produce joint-angle estimates within error margins sufficient for training and feedback applications. 
If successful, this approach could provide a practical means of supporting kinematic assessment outside laboratory settings, with potential applications in everyday golf practice and coaching.

Therefore, this paper proposes a deep learning framework for estimating full-body joint angles during the golf swing from a wrist-worn smartwatch IMU. The framework incorporates OMC-free IMU calibration and IMU-based swing segmentation. The framework combines IMU embedding with feature-wise linear modulation (FiLM)~\cite{perez2018film} and temporal convolutional encoding to learn wrist-to-body motion relationships. The framework is trained and evaluated using synchronized smartwatch inertial signals and ground truth kinematics derived from marker-based optical motion capture. The dataset includes thirty-six golfers across two skill levels, seven club types, and three swing-amplitude conditions. Using subject-wise cross-validation, we evaluate full-body joint-angle accuracy, trunk--pelvis rotational metrics, and the effects of skill level, club type, and swing amplitude on estimation error. OMC-assisted IMU calibration is additionally evaluated as a reference.

Our contributions can be summarized as follows: 

(i) We present a deep-learning framework for estimating full-body joint angles and golf-specific kinematic metrics during a golf swing using a wrist-worn 9-axis IMU, with OMC-free IMU calibration;

(ii) We present a comprehensive golf swing dataset of synchronized IMU data and OMC-derived full-body joint angles and joint centers, encompassing two skill levels, seven club types, and three swing-amplitude conditions to reflect real-world swing variability;

(iii) To the best of our knowledge, this is the first study to estimate full-body kinematics during golf swings from a real single wrist-worn IMU, with validation against an optical motion capture system, providing a basis for future biomechanical analysis in more practical golf settings.

\section{Related Work}
\label{sec:related_work}
\subsection{Laboratory-Based Golf Swing Kinematics}
Laboratory-based golf biomechanics studies have sought to identify biomechanical metrics associated with golf performance and skill discrimination. Hume et~al. summarized that effective golf performance depends on coordinated weight transfer, proximal-to-distal sequencing, trunk rotation, and appropriately timed wrist release, which together contribute to clubhead speed and shot distance~\cite{hume2005role}. 
Among trunk--pelvis variables, torso–pelvis separation (X-factor) is particularly important, as greater rotational separation between the trunk and pelvis stretches the connecting musculature to store elastic energy, which is subsequently released during the downswing to increase clubhead speed.
Myers et~al. reported that ball velocity was associated with X-factor and upper-torso rotational velocity, while Chu et~al. further identified X-factor, delayed release, trunk tilting, and weight shifting as significant contributors to driving ball velocity~\cite{myers2008role, chu2010relationship}.
Rotational metrics have also been used to characterize skill-related swing differences. Meister et~al. evaluated upper-torso rotation, pelvic rotation, X-factor, upper torso obliquity (S-factor), and pelvic obliquity (O-factor), and showed that professional benchmark curves based on these rotational metrics can reveal deviations in amateur golfers~\cite{meister2011rotational}. Zhou et~al. further demonstrated that pelvic and upper-torso rotational velocity metrics can be integrated into a swing performance index to distinguish professional from amateur golfers~\cite{zhou2022swing}.

\begin{figure*}[!t]
\centerline{\includegraphics[width=0.8\textwidth]{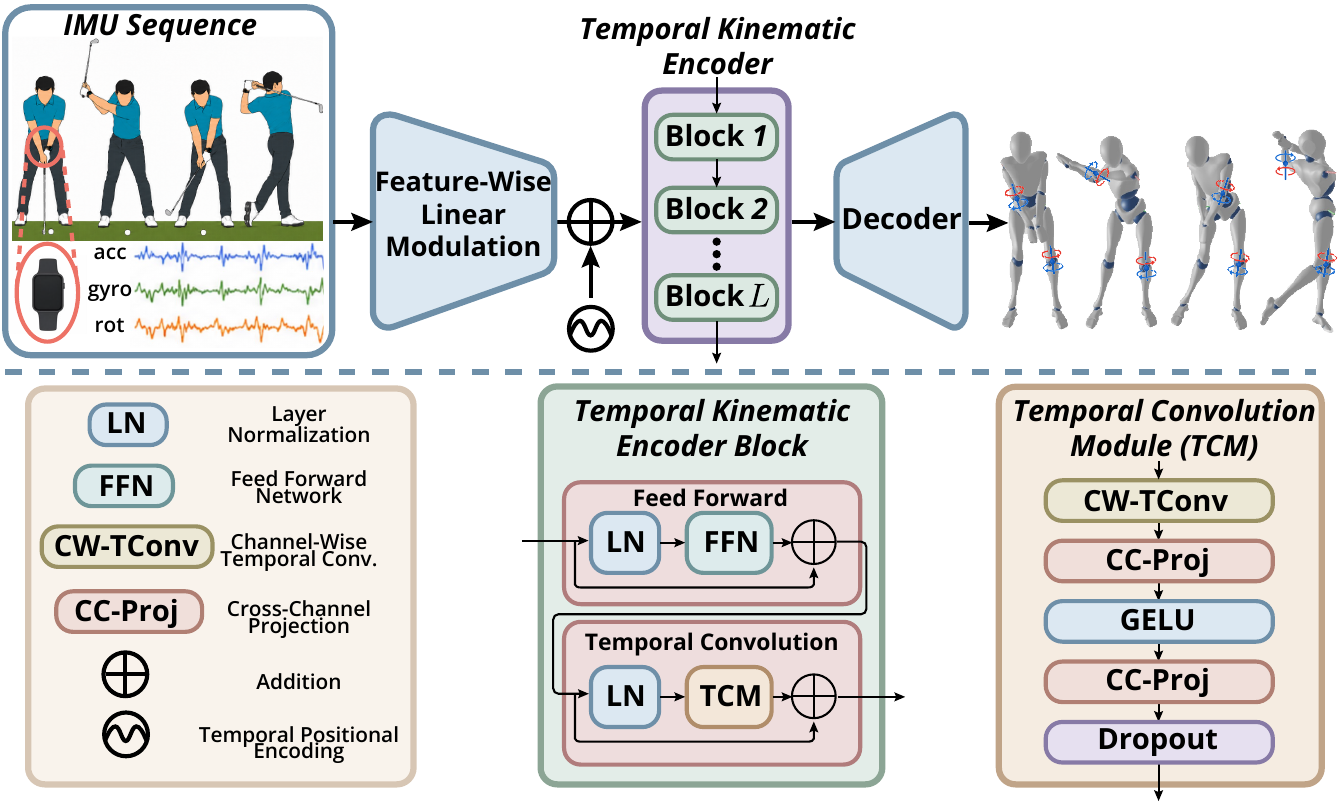}}
\caption{Architecture of the proposed Wrist-IMU Temporal Kinematic Network (WIT-KinNet). The input is a wrist-worn IMU sequence over $T$ frames, consisting of acceleration, angular velocity, and orientation signals. The three modalities are concatenated and embedded by a multilayer perceptron (MLP), followed by feature-wise linear modulation (FiLM) conditioned on orientation and locally averaged acceleration and angular-velocity magnitudes, and then augmented by adding temporal positional encoding. The resulting sequence is processed by $L$ temporal kinematic encoder blocks followed by an MLP decoder. Each temporal kinematic encoder block contains residual feed-forward and temporal convolution modules. The temporal convolution module combines channel-wise temporal convolution with two cross-channel projections to model local temporal dynamics. The MLP decoder maps the encoded temporal features to full-body joint-angle estimates.}
\label{fig:model}
\end{figure*}

\subsection{Vision-Based Golf Swing Analysis}
Vision-based methods aim to improve the practicality of golf swing analysis by estimating swing information from images or videos without requiring laboratory motion-capture systems. McNally et~al. introduced GolfDB, a video database for golf swing sequencing, and proposed SwingNet for golf swing event detection, achieving 76.1\% average event-detection accuracy across eight manually annotated swing events~\cite{mcnally2019golfdb}. Ju et~al. proposed GolfMate, which provides interpretable self-training feedback by comparing learner and professional poses~\cite{ju2023golfmate}. Lee et~al. proposed GolfPose, a golf-specific pose-estimation framework trained with Vicon-derived keypoint annotations, reducing the mean three-dimensional golfer keypoint error from 109.4 mm to 35.6 mm~\cite{lee2024golfpose}. Although vision-based methods improve accessibility, their reliance on fixed camera setups and recording typically restricts their use to driving ranges or indoor settings, limiting practical application during actual on-course play.

\subsection{Wearable IMU-Based Golf Swing Analysis}
IMU signals have been used for golf swing segmentation and wrist-trajectory tracking. Kim and Park used a single IMU placed on the wrist, head, or waist to segment the golf swing into major phases~\cite{kim2020golf}, achieving average segmentation errors of 5--92~ms across phase transition points. The same authors also used wrist-worn inertial sensors to track wrist trajectories during the golf swing, where sensor orientation was estimated from address-phase IMU signals and drift was mitigated using kinematic constraints~\cite{kim2024enhancing}.
IMUs have also been used to estimate selected joint- or segment-level biomechanical variables. Vo et~al. developed a multisensor smart glove for real-time wrist angle analysis and abnormal motion detection during the golf swing~\cite{vo2025multisensor}. 
Kim et~al. validated IMU-derived golf rotational biomechanics against motion capture, including upper-torso rotation, pelvic rotation, pelvic rotational velocity, S-factor, O-factor, and X-factor, using IMUs mounted directly on the T1 and L4 vertebrae~\cite{kim2023validation}. They reported strong agreement between IMU and motion capture measurements, with intraclass correlation coefficients~\cite{shrout1979intraclass} ranging from 0.91 for O-factor to 1.00 for upper-torso rotation and absolute mean differences in Bland--Altman analysis~\cite{bland1986statistical} ranging from 0.61$^\circ$ to 1.67$^\circ$ across parameters. While this segment-level placement likely contributed to the high measurement fidelity, by directly capturing torso and pelvic orientation rather than inferring it from a distal sensor, it also requires golfers to wear multiple dedicated sensors affixed to the trunk and pelvis, which is less practical for everyday on-course adoption than a single commercially available wrist-worn device such as a smartwatch.
More recently, Lauer explored single-wrist inertial sensing for full-body golf swing analysis using a virtual wrist-worn inertial sensor~\cite{lauer2025learning}. However, the inertial inputs and reference kinematics were generated from a video-based reconstruction pipeline~\cite{shin2024wham}, leaving OMC-validated full-body golf swing reconstruction from real wrist-worn smartwatch IMU measurements unresolved.

\section{Methods}
\label{sec:Methods}
We propose the Wrist-IMU Temporal Kinematic Network (WIT-KinNet), a single-IMU framework for estimating full-body joint angles during golf swings from a wrist-worn IMU (Fig.~\ref{fig:model}). WIT-KinNet employs IMU embedding with feature-wise linear modulation (FiLM), temporal convolutional encoding, and a multilayer perceptron (MLP) decoder to estimate frame-wise full-body joint rotations.

\subsection{Problem Formulation}

Given a sequence of wrist-worn IMU signals recorded during a golf swing, our objective is to learn a mapping function from the input IMU sequence to the corresponding full-body joint rotations:
\begin{equation}
    f: \mathcal{X} \rightarrow \hat{\mathcal{Y}},
\end{equation}
where $\mathcal{X}=\{\mathbf{x}_{t}\}_{t=1}^{T}\in\mathbb{R}^{T\times 12}$ denotes the input IMU sequence over $T$ frames, with each frame represented by $\mathbf{x}_{t}=[\tilde{\mathbf{a}}_{t},\boldsymbol{\omega}_{t},\mathbf{o}_{t}]$. 
Here, $\tilde{\mathbf{a}}_{t}\in\mathbb{R}^{3}$ and $\boldsymbol{\omega}_{t}\in\mathbb{R}^{3}$ denote gravity-compensated smartwatch linear acceleration and angular velocity expressed in the reference frame $F$ defined in Section~\ref{sec:Data_Processing}, respectively, while $\mathbf{o}_{t}\in\mathbb{R}^{6}$ is the 6D representation~\cite{zhou2019continuity} of the forearm orientation relative to the reference frame $F$.
$\hat{\mathcal{Y}}=\{\hat{\mathbf{y}}^{j}_{t}\mid t=1,\ldots,T,\ j\in\mathcal{A}\}\in\mathbb{R}^{T\times|\mathcal{A}|\times 6}$ denotes the predicted joint-rotation sequence, where $\hat{\mathbf{y}}^{j}_{t}\in\mathbb{R}^{6}$ is the estimated 6D rotation of joint $j$ at frame $t$, and $\mathcal{A}$ denotes the target joint set. The 6D rotation representation is used to provide a continuous rotation parameterization and avoid gimbal-lock issues associated with Euler-angle outputs.


\subsection{Wrist-IMU Temporal Kinematic Network}
\label{subsec:singleimu_transformer}

The proposed Wrist-IMU Temporal Kinematic Network (WIT-KinNet) estimates full-body golf swing joint rotations from the preprocessed wrist-IMU sequence. It consists of three components: IMU embedding with FiLM, temporal sequence modeling, and an MLP decoder.

\subsubsection{IMU Embedding and Feature-Wise Linear Modulation}
WIT-KinNet jointly embeds the concatenated modalities and applies FiLM to condition the resulting features on forearm orientation and local motion magnitude:
\begin{subequations}
\label{eq:imu_embedding}
\begin{align}
    \mathbf{u}_{t}
    &=
    f_{\mathrm{imu}}(\mathbf{x}_{t}),\\
    \mathbf{c}_{t}
    &=
    [\mathbf{o}_{t},m_{a,t},m_{\omega,t}],\\
    [\boldsymbol{\gamma}_{t},\boldsymbol{\beta}_{t}]
    &=
    f_{\mathrm{FiLM}}(\mathbf{c}_{t}),\\
    \mathbf{z}^{0}_{t}
    &=
    \bigl(1+\tanh(\boldsymbol{\gamma}_{t})\bigr)
    \odot\mathbf{u}_{t}
    +\boldsymbol{\beta}_{t}
    +\mathbf{p}_{t}.
\end{align}
\end{subequations}
Here, $f_{\mathrm{imu}}$ is a shared two-layer MLP with layer normalization, GELU activation, and dropout between its linear layers. The scalars $m_{a,t}$ and $m_{\omega,t}$ are the mean Euclidean magnitudes of the acceleration and angular velocity within a centered temporal window. Together with the 6D orientation, they form the conditioning vector $\mathbf{c}_{t}\in\mathbb{R}^{8}$.
The conditioning MLP $f_{\mathrm{FiLM}}$ maps $\mathbf{c}_{t}$ to channel-wise scaling and shifting parameters $\boldsymbol{\gamma}_{t},\boldsymbol{\beta}_{t}\in\mathbb{R}^{d}$, and $\odot$ denotes element-wise multiplication. This modulation allows the embedded features to adapt to orientation and motion magnitude~\cite{perez2018film}. The vector $\mathbf{p}_{t}$  denotes sinusoidal temporal positional encoding.

The embedded sequence is denoted as $\mathbf{Z}^{0}\in\mathbb{R}^{B\times T\times d}$, where $B$, $T$, and $d$ are the batch size, sequence length, and latent feature dimension, respectively.

\subsubsection{Temporal Sequence Modeling}

Since a wrist-worn IMU provides only sparse observations of whole-body motion, temporal context is critical for estimating full-body joint rotations. Each encoder block refines the embedded representation through a frame-wise feed-forward network and a temporal convolution module (TCM). For the $\ell$-th encoder block, where $\ell=0,1,\ldots,L-1$, the feature sequence is updated by
\begin{subequations}
\label{eq:temporal_encoder}
\begin{align}
    \bar{\mathbf{Z}}^{\ell}
    &=
    \mathbf{Z}^{\ell}
    +
    \operatorname{FFN}
    \left(
    \operatorname{LN}(\mathbf{Z}^{\ell})
    \right),\\
    \mathbf{Z}^{\ell+1}
    &=
    \bar{\mathbf{Z}}^{\ell}
    +
    \operatorname{TCM}
    \left(
    \operatorname{LN}(\bar{\mathbf{Z}}^{\ell})
    \right).
\end{align}
\end{subequations}
The temporal convolution module is defined as
\begin{equation}
\begin{aligned}
    \operatorname{TCM}(\mathbf{U})
    =
    \operatorname{Drop}\Bigl(
    \operatorname{P}_{2}\bigl(
    \operatorname{GELU}\bigl(
    \operatorname{P}_{1}(
    \operatorname{Conv}_{K}(\mathbf{U}))
    \bigr)\bigr)\Bigr).
\end{aligned}
\end{equation}
where $\mathbf{Z}^{\ell}$ denotes the input to the $\ell$-th encoder block, and $\bar{\mathbf{Z}}^{\ell}$ denotes the intermediate output after the feed-forward submodule. $L$ is the total number of temporal kinematic encoder blocks. $\operatorname{LN}(\cdot)$, $\operatorname{FFN}(\cdot)$, and $\operatorname{GELU}(\cdot)$ denote layer normalization, a frame-wise feed-forward network, and the Gaussian error linear unit activation function, respectively. $\operatorname{Conv}_{K}(\cdot)$ denotes a depthwise temporal convolution with kernel size $K$, while $\operatorname{P}_{1}(\cdot)$ and $\operatorname{P}_{2}(\cdot)$ denote the first and second cross-channel projection layers in the temporal convolution module, respectively.


\subsubsection{MLP Decoder}

After temporal sequence modeling, the final contextual representation $\mathbf{Z}^{L}$ is passed through layer normalization and decoded by a frame-wise MLP decoder shared across all time steps. The decoder maps each $d$-dimensional temporal feature to the 6D rotations of all target joints, producing the predicted joint-rotation sequence $\hat{\mathbf{Y}}\in\mathbb{R}^{B\times T\times|\mathcal{A}|\times6}$, where $B$ is the batch size, $T$ is the sequence length, and $\mathcal{A}$ denotes the target joint set.


\begin{figure*}[!t]
    \centering
    \subfloat[]{
        \label{fig:markers}
        \includegraphics[height=.3\textheight]{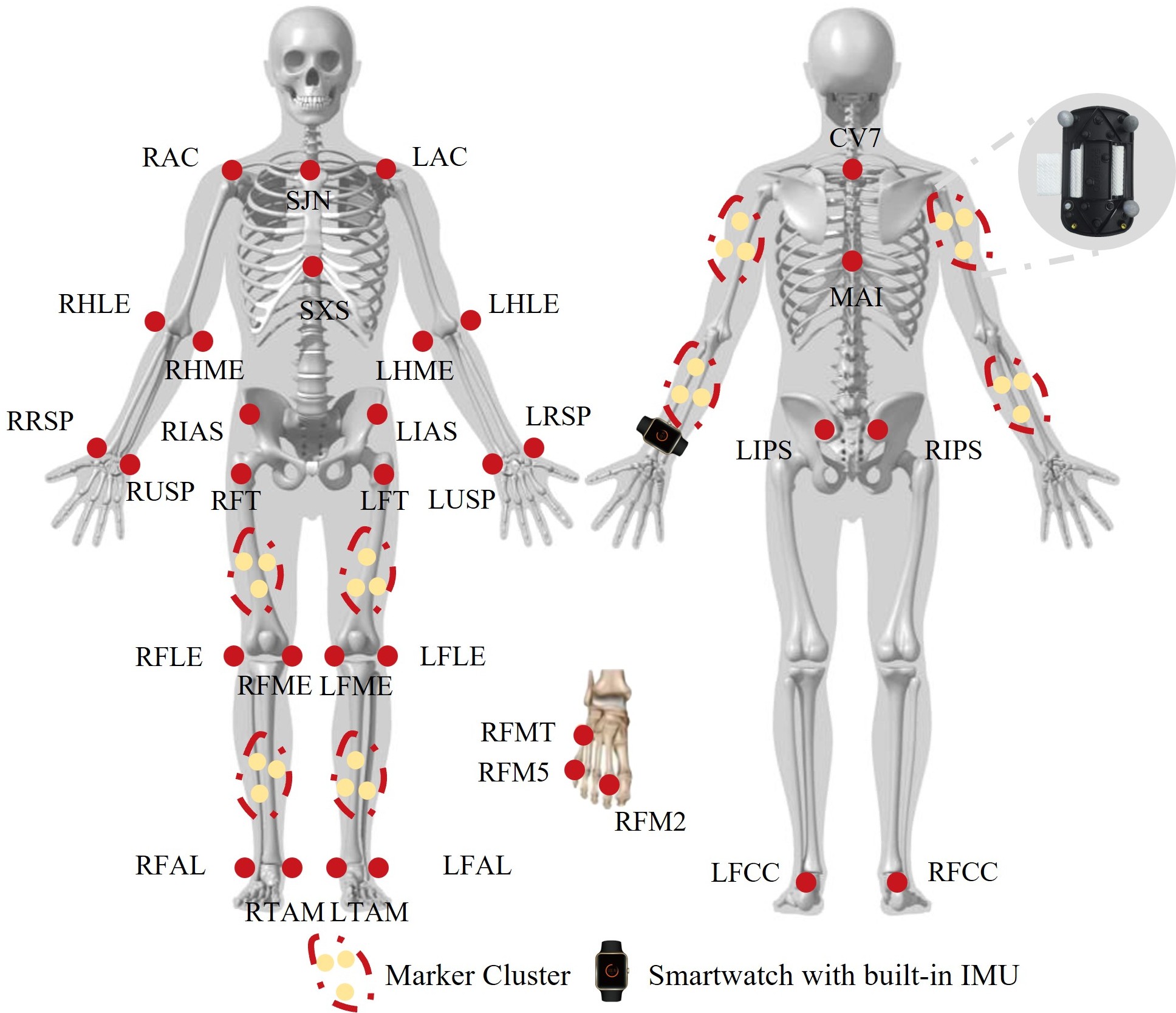}
    }
    \subfloat[]{
        \label{fig:screenshot}
        \includegraphics[height=.3\textheight]{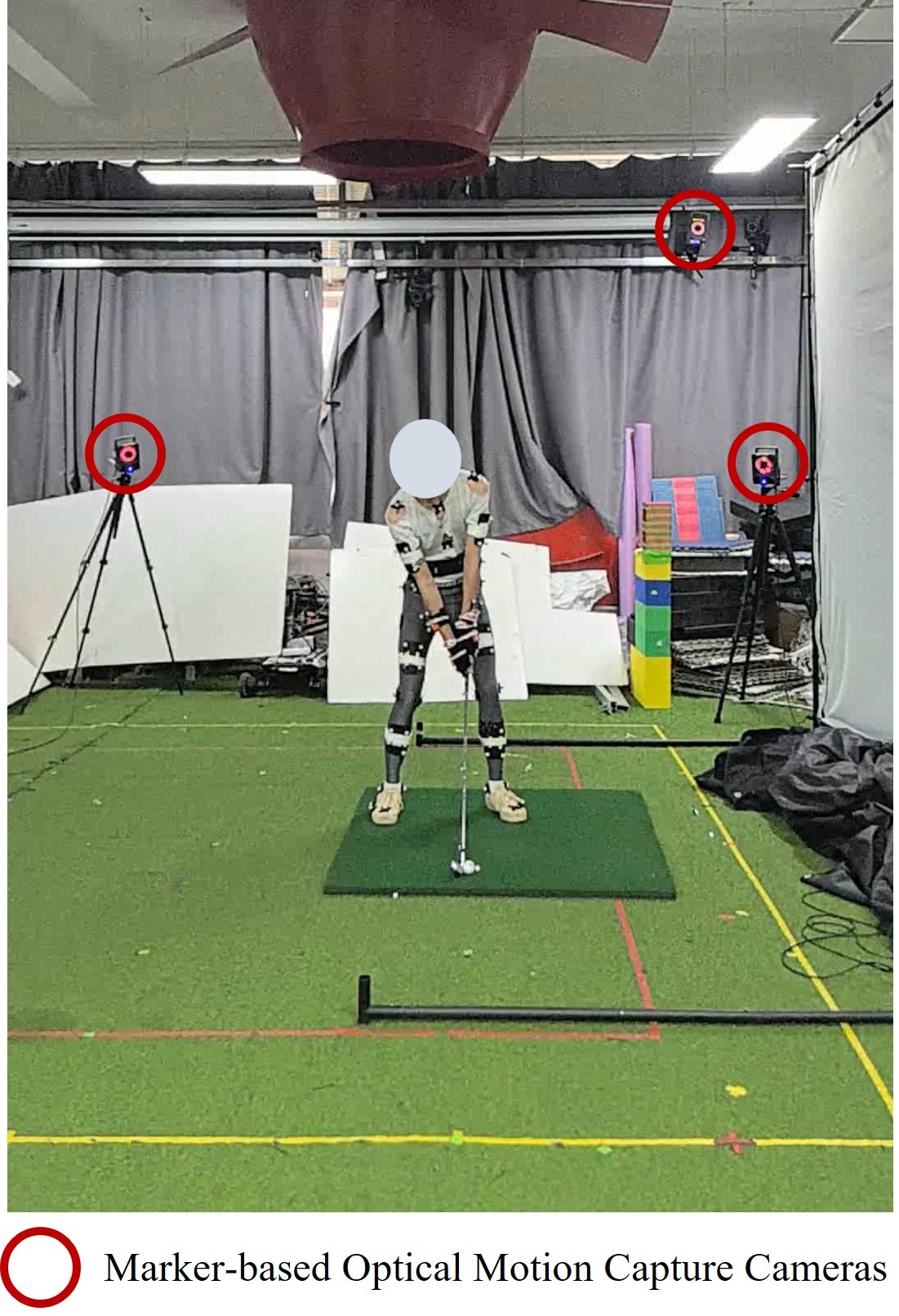}
    }
    \caption{Experimental configuration.
    (a) Placement of anatomical markers, marker clusters, and the wrist-worn smartwatch IMU.
    (b) Laboratory data collection setup during a golf swing trial. Participants hit golf balls into a hitting screen positioned along the intended ball flight direction, while optical motion capture cameras tracked marker trajectories to compute ground-truth kinematics.}
    \label{fig:setup}
\end{figure*}


\subsection{Training Objective}

The model was trained using rotation reconstruction and first- and second-order temporal consistency losses~\cite{zhu2023motionbert, zeng2022smoothnet}. Let $\mathbf{y}_{btj}\in\mathbb{R}^{6}$ and $\mathbf{R}_{btj}$ denote the reference 6D rotation and corresponding rotation matrix of joint $j$ at frame $t$ in sequence $b$. Setting $\mathbf{D}^{(0)}_{btj}=\mathbf{R}_{btj}$, we define
\begin{equation}
\mathbf{D}^{(k)}_{btj}
=(\mathbf{D}^{(k-1)}_{btj})^{T}
\mathbf{D}^{(k-1)}_{b,t+1,j},
\quad k=1,2,
\end{equation}
with predicted quantities defined analogously. The three error terms are
\begin{equation}
\begin{aligned}
\ell^{(0)}_{btj}
&=\|\hat{\mathbf{y}}_{btj}-\mathbf{y}_{btj}\|_2^2,\\
\ell^{(k)}_{btj}
&=\|\hat{\mathbf{D}}^{(k)}_{btj}-\mathbf{D}^{(k)}_{btj}\|_F^2,
\quad k=1,2.
\end{aligned}
\end{equation}
The training objective is
\begin{equation}
\mathcal{L}
=\sum_{k=0}^{2}\frac{\lambda_k}{N_k}
\sum_{b=1}^{B}\sum_{t=1}^{T_b-k}\sum_{j=1}^{J}
\ell^{(k)}_{btj},
\end{equation}
where $B$ is the batch size, $J$ is the number of predicted rotations, $T_b$ is the sequence length before padding, and $N_k=J\sum_{b=1}^{B}(T_b-k)$. Thus, padded positions are excluded from the objective. The coefficients $\lambda_k$ weight rotation reconstruction and first- and second-order temporal consistency, respectively; $\|\cdot\|_2$ and $\|\cdot\|_F$ denote the Euclidean and Frobenius norms.
\section{Experimental Validation}
\label{sec:Experimental_Validation}

\subsection{Experiment Setup}
\subsubsection{Subjects}
Thirty-six right-handed subjects participated in the experiment, including fourteen beginner golfers and twenty-two skilled golfers (Table~\ref{tab:subjects}). Beginners had standard course handicap greater than 22. Skilled golfers had a handicap of 22 or lower. The study protocol adhered to the principles of the Declaration of Helsinki and was approved by the Ethics Committee of Shanghai Jiao Tong University (No.~E2021013P).

\begin{table}[!t]
\centering
\caption{Participant Demographics and Golf Experience}
\label{tab:subjects}
\scriptsize
\setlength{\tabcolsep}{2.5pt}
\renewcommand{\arraystretch}{1.10}
\begin{tabular}{lcccccc}
\hline
Group & $n$ & Sex & Age & Height & Weight & Experience \\
      &     & (M/F) & (years) & (m) & (kg) & (years) \\
\hline
Beginner      
& 14 & 10/4 & $ 28.3\pm 8.4$ & $1.72 \pm 0.08$ 
& $69.9 \pm 13.8$ & $0.9 \pm 0.6$ \\

Skilled
& 22 & 16/6 & $27.2 \pm 9.6$ & $1.75 \pm 0.08$ 
& $68.0 \pm 11.3$ & $7.5 \pm 6.1$ \\

\hline
All subjects  
& 36 & 26/10 & $27.6 \pm 9.0$ & $1.74 \pm 0.08$ 
& $68.7 \pm 12.2$ & $5.0 \pm 5.8$ \\
\hline
\end{tabular}
\end{table}

\subsubsection{Marker Placement}
Thirty-six reflective markers (14 mm) were placed on anatomical landmarks for segment definition: left and right acromion (LAC/RAC), sternum jugular notch (SJN), sternum xiphisternal joint (SXS), seventh cervical vertebra (CV7), midway between the inferior angles of most caudal points of the two scapula (MAI), lateral and medial humerus epicondyles (LHLE/RHLE/LHME/RHME), radius-styloid and ulna-styloid processes (LRSP/RRSP/LUSP/RUSP), bilateral anterior superior iliac spines (LIAS/RIAS), posterior superior iliac spines (LIPS/RIPS), femur greater trochanter (LFT/RFT), lateral and medial femoral epicondyles (LFLE/RFLE/LFME/RFME), fibula apex of lateral and tibia apex of medial malleoli (LFAL/RFAL/LTAM/RTAM), head of the second metatarsal (LFM2/RFM2), head of the fifth metatarsal (LFM5/RFM5), tuberosity of the fifth metatarsal (LFMT/RFMT), and posterior surface of the calcaneus (LFCC/RFCC). Eight clusters, each with three markers on its surface, were placed on the left and right upper arms, forearms, thighs, and shanks, respectively, for segment tracking during motion capture (Fig.~\ref{fig:markers}). Marker trajectories were recorded at 120~Hz using a nine-camera OMC system (Vicon, Oxford Metrics Group, Oxford, U.K.) (Fig.~\ref{fig:screenshot}).

\subsubsection{IMU Placement}
A commercial smartwatch (Huawei Watch GT 5 Pro, Huawei Technologies Co., Ltd., Shenzhen, China) with an embedded nine-axis IMU was worn on the left (lead) wrist (Fig.~\ref{fig:markers}). The IMU recorded three-axis acceleration and angular velocity signals at 100~Hz and three-axis magnetometer signals at 10~Hz. For each axis, the accelerometer had a measurement range of $\pm 8g$ with a resolution of $0.000244g$, where $g$ denotes gravitational acceleration, while the gyroscope had a measurement range of $\pm 40.0~\mathrm{rad/s}$ with a resolution of $0.00122~\mathrm{rad/s}$.

\subsection{Experiment Procedure}
Retroreflective markers were attached and the IMU device was secured to the subject. Subjects then performed figure-eight motions with the smartwatch IMU for thirty seconds to calibrate magnetometer hard-iron and soft-iron distortions.
Each subject performed golf swings striking the ball into a hitting screen using seven club types  (Fig.~\ref{fig:screenshot}), including a driver (1W), 3-wood (3W), 5-hybrid (5H), 5-iron (5I), 7-iron (7I), 9-iron (9I), and sand wedge (SW). For each club, subjects performed five swings at each of three effort levels: full, half, and quarter.
For full swings, subjects were instructed to perform their natural swing using maximal comfortable range of motion. For half and quarter swings, subjects were instructed to stop their backswing when either the lead arm approached horizontal, or reached thorax height, respectively.  
Each trial consisted of five golf swings performed using a specific club under a specific swing amplitude. At the beginning of each trial, subjects performed a five-second static T-pose, followed by three consecutive left-arm swings for further data synchronization during data processing.

\subsection{Data Processing}
\label{sec:Data_Processing}

The marker trajectories captured by the OMC system were processed in Visual3D (C-Motion, MD, USA) to obtain ground-truth joint angles following the CAST procedure~\cite{cappozzo1995position}. 
Joint angles were subsequently derived from the relative orientation between adjacent segments using a flexion-abduction-rotation Euler angle sequence. The joints included bilateral hip, knee, ankle, shoulder, and elbow joint angles, as well as lumbar angles and pelvis. 
Joint angles were transformed into the 6D rotation representation~\cite{zhou2019continuity} to provide a continuous and stable rotation parameterization for deep learning-based modeling.

The accelerometer and gyroscope signals from the IMUs were linearly interpolated from 100~Hz to a common reference grid at 120~Hz to match the sampling rate of the OMC system. The magnetometer measurements were first corrected for hard- and soft-iron distortions using ellipsoid fitting. The corrected magnetic-field vectors were resampled from 10~Hz to one third of the reference-grid rate, 40~Hz, using spherical linear interpolation of their unit directions and linear interpolation of their magnitudes. The IMU orientations were then estimated using an open-source complementary filtering algorithm~\cite{fan2018improving}, with gyroscope and accelerometer signals processed at the reference-grid rate and magnetic feedback applied every third update. OMC and IMU data were synchronized by cross-correlating angular velocity magnitudes from the left forearm marker cluster and IMU during the initial three left-arm swings. Each trial was then manually divided into five recording segments to separate individual swings, without requiring the segment boundaries to coincide with address or finish.

Two IMU calibration procedures were considered: OMC-free calibration as the primary approach and OMC-assisted calibration as a laboratory reference. A static T-pose was recorded at the beginning of each trial and used for both calibration methods. During this pose, subjects faced the negative $X$ direction of the OMC coordinate system, with their left arms parallel to the intended ball-flight direction. A reference frame was denoted by \(F\). For OMC-assisted calibration, \(F\) coincided with the OMC coordinate frame \(O\). For OMC-free calibration, its \(X\)-, \(Y\)-, and \(Z\)-axes pointed forward, leftward, and vertically upward, respectively, relative to the static T-pose. Let \(I\), \(C\), and \(S\) denote the IMU global frame, smartwatch carrier frame, and forearm segment frame, respectively.

For OMC-free calibration, subjects assumed a static T-pose with their
left arms parallel to the intended ball-flight direction. Correct smartwatch placement was assumed, with a fixed carrier-to-segment rotation \({}^{S}\mathbf{R}_{C}\) specified by the axis ordering and sign conventions of the two coordinate systems. Mean acceleration and magnetic-field measurements during the static T-pose were used to estimate the carrier orientation in frame \(I\). The horizontal projection of the smartwatch \(+X\)-axis, directed from the left elbow toward the wrist, defined the leftward direction. Together with the vertical direction, this determined the reference-frame alignment \({}^{F}\mathbf{R}_{I}\).

For OMC-assisted calibration, subjects faced the negative \(X\)-axis of the OMC frame during the static T-pose. Since the OMC and IMU global frames shared the same vertical direction, \({}^{F}\mathbf{R}_{I}\) was obtained by aligning the IMU-derived heading with the known subject-facing direction. Using the forearm orientation \({}^{F}\mathbf{R}_{S}(t_{0})\) obtained from Visual3D at the static calibration pose \(t_{0}\), the carrier-to-segment rotation was determined as
\begin{equation}
{}^{S}\mathbf{R}_{C}
=
\left({}^{F}\mathbf{R}_{S}(t_{0})\right)^{T}
{}^{F}\mathbf{R}_{I}
{}^{I}\mathbf{R}_{C}(t_{0}).
\end{equation}

For either procedure, the forearm orientation in the corresponding reference frame was obtained as
\begin{equation}
{}^{F}\mathbf{R}_{S}(t)
=
{}^{F}\mathbf{R}_{I}
{}^{I}\mathbf{R}_{C}(t)
\left({}^{S}\mathbf{R}_{C}\right)^{T}.
\end{equation}
The carrier-frame accelerometer specific force \({}^{C}\mathbf{a}(t)\) and angular velocity \({}^{C}\boldsymbol{\omega}(t)\) were transformed as
\begin{equation}
{}^{F}\mathbf{a}(t)
=
{}^{F}\mathbf{R}_{I}
{}^{I}\mathbf{R}_{C}(t)
{}^{C}\mathbf{a}(t),
\end{equation}
\begin{equation}
{}^{F}\boldsymbol{\omega}(t)
=
{}^{F}\mathbf{R}_{I}
{}^{I}\mathbf{R}_{C}(t)
{}^{C}\boldsymbol{\omega}(t).
\end{equation}
The linear acceleration input was obtained as
${}^{F}\tilde{\mathbf{a}}_{t}={}^{F}\mathbf{a}(t)-{}^{F}\mathbf{g}$,
where ${}^{F}\mathbf{g}$ denotes the gravity component of the
accelerometer measurements, expressed in frame $F$.

The forearm orientations were further converted into the 6D rotation representation.
For ground-truth preparation, the global pelvis orientation was expressed in the corresponding reference frame, while relative joint rotations remained unchanged. Sensor quality statistics and input sensitivity analyses are provided in Sections~IV and~V of the Supplementary Material, respectively.

Each swing cycle was divided into three event-aligned phases: backswing (address to top of backswing), downswing (top of backswing to ball-impact), and follow-through (ball-impact to end of swing).
Ground truth phase timestamps were obtained automatically by a simple marker-based event detection algorithm applied to marker trajectories captured by the optical motion capture system. 
An auxiliary bidirectional long short-term memory (BiLSTM) network based on the architecture described by Kim and Park~\cite{kim2020golf} was used to estimate four swing event times (address, top of backswing, ball-impact, and end of swing) from smartwatch acceleration and angular velocity signals.

\subsection{Implementation and Training Details}

The proposed network was implemented in Python 3.10 using PyTorch 2.8.0 and trained on a single NVIDIA GeForce RTX 4070 Ti Super GPU. 
The latent feature dimension is set to $d=128$, and the temporal encoder comprises $L=2$ stacked residual blocks. The IMU embedding MLP has input, hidden, and output dimensions of 12, 128, and 128, respectively. The corresponding dimensions of the FiLM conditioning MLP are 8, 32, and 256. Each FFN uses 512 hidden units, with input and output dimensions of 128.
The channel-wise temporal convolution employs a kernel size of $K=5$. Its two pointwise projections expand the channel dimension from 128 to 256 and then reduce it to 128. The decoder maps 128-dimensional features through two hidden layers of 256 units each to 72 outputs, producing 12 joint rotations in the 6D representation. Dropout was set to 0.1. Predicted 6D rotations were converted to rotation matrices by Gram--Schmidt orthogonalization using the row-vector convention, and then to intrinsic flexion-abduction-rotation Euler angles.
Since swing lengths vary within a batch, the sequence length $T$ is set to the maximum sequence length in each batch, and a binary padding mask is applied to exclude padded frames from all computations. For the training objective, the loss weighting coefficients were set to $\lambda_{\mathrm{0}}=1.25$, $\lambda_{\mathrm{1}}=0.2$, and $\lambda_{\mathrm{2}}=0.05$ for rotation reconstruction and first- and second-order temporal consistency, respectively.
For golf swing phase segmentation, the BiLSTM network used three-axis acceleration, three-axis angular velocity, and normalized time as inputs, with five bidirectional layers containing 32 hidden units per direction and a five-class frame-wise output layer. It was trained using a weighted negative log-likelihood of the summed probabilities of phases consistent with the OMC-derived ground truth event times. The proposed model, all baseline models, and the BiLSTM segmentation network were optimized using the AdamW~\cite{loshchilov2017decoupled} optimizer  with a weight decay of $0.01$, a learning rate of $2\times10^{-4}$, and a batch size of $4$ over $50$ training epochs. All models were trained using the same random seed, and the weights from each model's final training epoch were used for testing.
Inference was evaluated on an NVIDIA GeForce RTX 4070 Ti SUPER GPU using FP32 and a batch size of one. The proposed pose estimation network and the segmentation BiLSTM contain 0.577 million and 0.111 million parameters, requiring 2.20 MiB and 0.42 MiB of parameter storage, respectively. Across all sequences, their mean forward latencies per sequence were 1.99 and 1.50 ms on the GPU, respectively.

\subsection{Baseline Comparisons and Ablation Studies}
The proposed model was compared with six baselines: a frame-wise multilayer perceptron (MLP), a bidirectional long short-term memory (BiLSTM) network~\cite{graves2005framewise}, a non-causal dilated temporal convolutional network (TCN)~\cite{bai2018empirical}, a vanilla Transformer~\cite{vaswani2017attention}, a Conformer-inspired convolution-augmented Transformer~\cite{gulati2020conformer}, and a network proposed by
Lauer~\cite{lauer2025learning}.
Ablation studies were conducted to assess the contributions of FiLM, TCM, and the three input modalities. Five variants were evaluated by separately removing FiLM, TCM, acceleration, angular velocity, or orientation while retaining the remaining model components. 
All baseline and ablation experiments used the same input preprocessing, training objectives, and optimization settings.

\subsection{Performance Evaluation}
The proposed Wrist-IMU Temporal Kinematic Network (WIT-KinNet) was evaluated using subject-wise five-fold cross-validation. To balance skill-level composition, participants of the two levels were assigned to the five folds as evenly as possible. Unless otherwise stated, all results were obtained using OMC-free calibration. The same IMU-based segmentation method was used for all conditions, including OMC-assisted calibration. The skill-level composition of each fold and IMU event-detection errors are reported in Sections~I and~VI of the Supplementary Material, respectively. Joint-angle estimation accuracy was quantified using the mean absolute error (MAE) for each predicted joint axis. Overall MAE was calculated within each swing by averaging bilateral
joint-angle MAEs and then the evaluated joint-angle components.
Overall, component-specific, and regional MAEs were first averaged
across swings within each subject, and their means and SDs were then
calculated across subjects. Boxplots show swing-level distributions. Comparisons of the proposed network with baseline models,
ablation variants, and the OMC-assisted calibration condition
used two-sided paired $t$-tests on subject-level mean MAEs,
with Holm correction~\cite{holm1979simple} within comparison
families. In addition to joint-level errors, five golf-specific kinematic variables closely associated with swing performance were further selected and evaluated: pelvic rotation, upper torso rotation, the difference between upper torso and pelvic rotation (X-factor), upper torso obliquity (S-factor), and pelvic obliquity (O-factor)~\cite{myers2008role, zhou2022swing, hume2005role, jung2022effects, meister2011rotational}. Each performance variable was computed from the predicted and ground truth joint rotations, zeroed at IMU-predicted address, and compared over an event-normalized swing cycle. 
Estimation errors of performance variables were also quantified using MAE, which was computed separately within each of the three phases and over the entire swing cycle. Whole-swing MAEs were computed on the original time grid, whereas phase-specific MAEs were computed from event-normalized profiles. MAE summaries were calculated across subject-specific means. Waveform plots show means and SDs across swings. Correlation between predicted and ground truth was quantified using Pearson's correlation coefficient ($r$), which was computed between the predicted and ground truth profile over the entire swing cycle within each swing and then aggregated using Fisher's transformation with equal subject weighting. 
Three trunk--pelvis rotational metrics, including pelvic rotation, upper torso rotation, and X-factor, were extracted at both their negative peak values and the IMU-predicted ball-impact event~\cite{meister2011rotational, myers2008role, cheetham2001importance, chu2010relationship}. Predicted and ground truth values for these metrics were compared using Pearson's correlation coefficient, two-way mixed-effects, single-measure absolute-agreement intraclass correlation coefficient (ICC)~\cite{shrout1979intraclass}, and Bland--Altman analysis~\cite{bland1986statistical}. 
Bland--Altman bias was estimated using a subject random-intercept model of predicted-minus-reference differences, and the limits of agreement were defined as the bias $\pm 1.96$ $\sqrt{\sigma_b^2+\sigma_w^2}$, incorporating between- and within-subject variance. Confidence intervals were obtained by subject-cluster bootstrap. To assess whether estimation accuracy was affected by swing condition, swing-level MAE was analyzed using linear mixed-effects models~\cite{laird1982random} for the five golf-specific performance variables. Skill level, club group, and swing amplitude were treated as fixed effects, with correlated subject-specific random intercepts and slopes for club group and swing amplitude. Club type was set as a two-level factor, with long clubs including 1W, 3W, and 5H, and irons/wedges including 5I, 7I, 9I, and SW.

\section{Results}
\label{sec:results}
\subsection{Overall Joint-Angle Accuracy}

\begin{figure}[!t]
\centerline{\includegraphics[width=0.95\columnwidth]{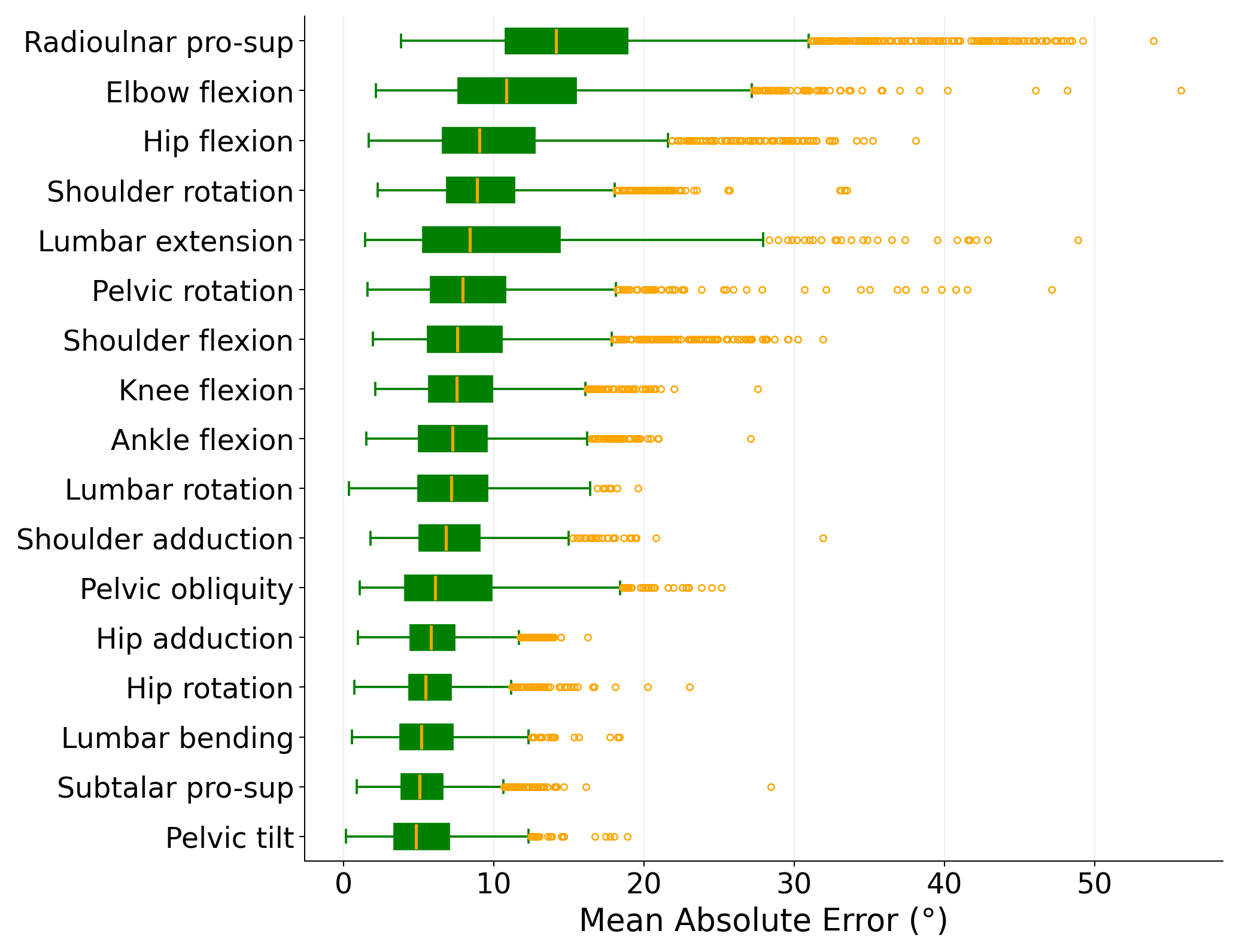}}

\caption{Distribution of joint-angle mean absolute errors across all swings. Bilateral joints were merged by averaging left and right errors within each swing. Boxes indicate the interquartile range, orange lines indicate medians, whiskers indicate non-outlier ranges, and orange circles indicate outlier swings.}

\label{fig:all-joint}
\end{figure}

\begin{figure}[!t]
\centerline{\includegraphics[width=0.9\columnwidth]{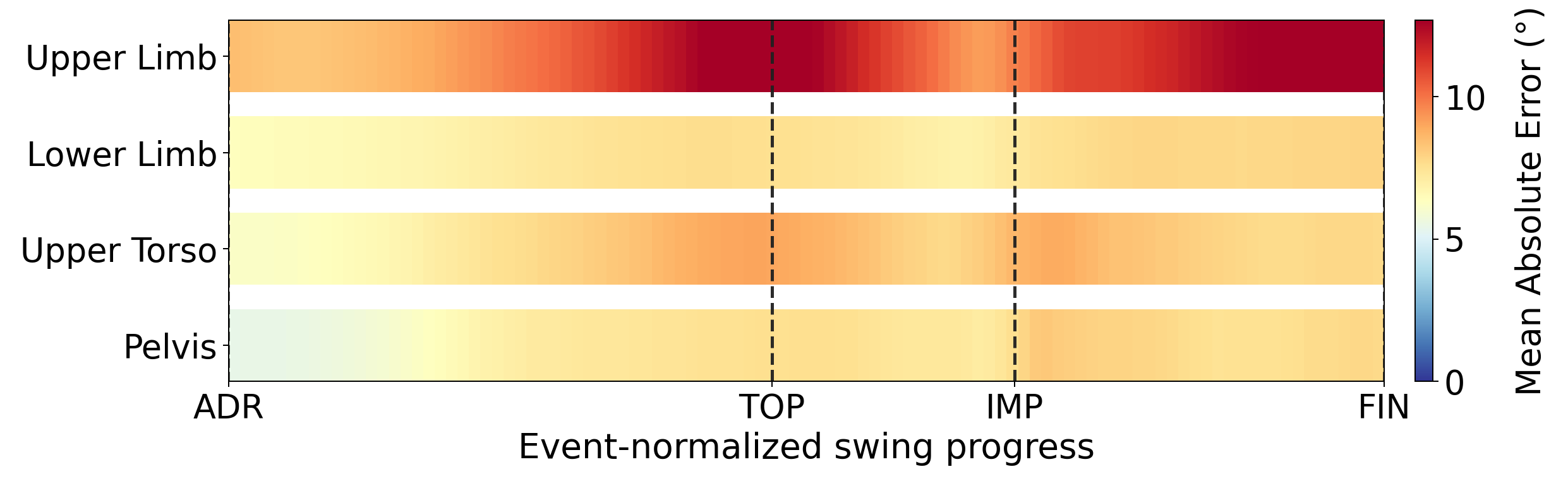}}
\caption{Event-aligned regional joint-angle mean absolute error across the golf swing. The swing cycle was aligned using four key events: address (ADR), top of backswing (TOP), ball-impact (IMP), and finish (FIN). Events were estimated from the IMU.} Upper limb, lower limb, upper torso, and pelvis errors were computed by averaging the joint-angle components within each region.

\label{fig:all-joint-phase}
\end{figure}

\begin{figure}[!t]
\centering
\includegraphics[width=0.95\columnwidth]{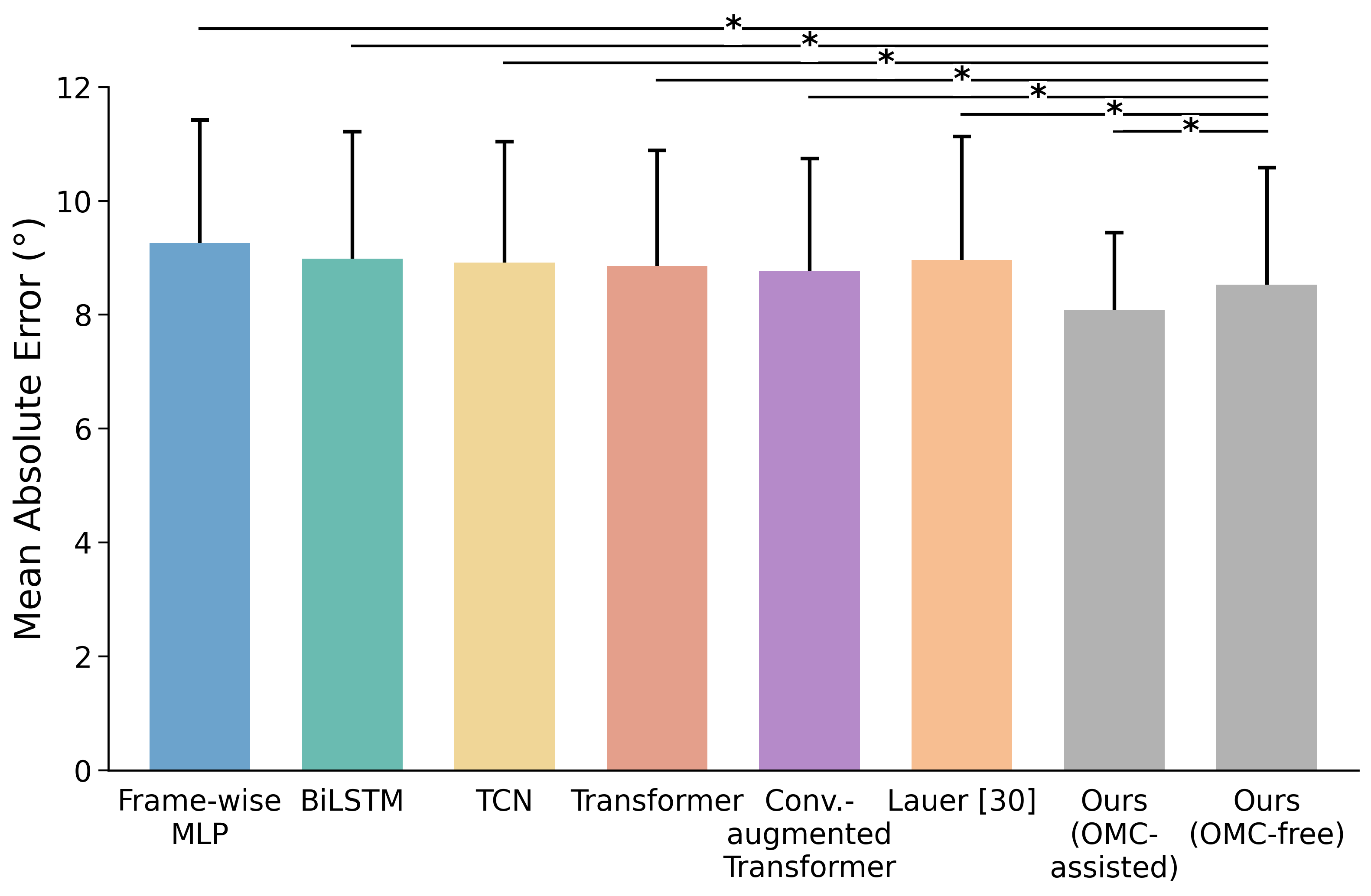}
\caption{Full-body joint-angle estimation accuracy of the proposed method, baseline models, and the proposed model with OMC-assisted calibration. Bars show the mean MAE across subjects, and error bars indicate the corresponding SD. Asterisks denote significant paired $t$-tests after Holm correction ($p<0.05$).}
\label{fig:baseline-comparison}
\end{figure}

\begin{figure}[!t]
\centering
\includegraphics[width=0.85\columnwidth]{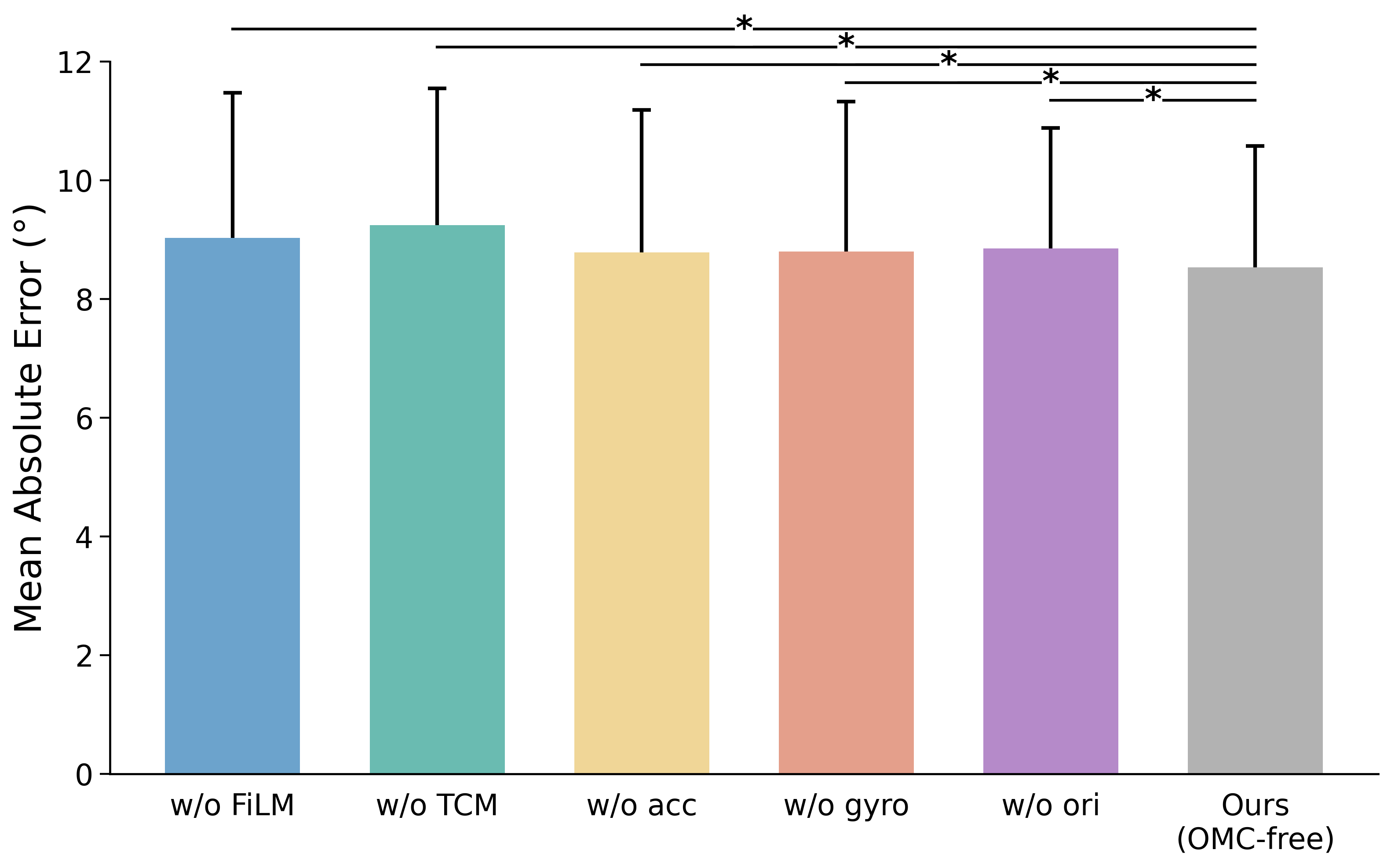}
\caption{Ablation study of model components and input modalities for full-body joint-angle estimation. Bars and error bars indicate the mean and SD across subject-specific mean MAEs. Asterisks denote significant paired $t$-tests after Holm correction ($p<0.05$).}
\label{fig:ablation-comparison}
\end{figure}

Across all evaluated joint-angle components, the proposed framework using OMC-free calibration achieved an overall MAE of $8.53\pm2.05^\circ$ (mean$\pm$SD; 95\% CI: $7.96$--$9.11^\circ$), with a median error of $8.22^\circ$ and an interquartile range of $7.13$--$9.30^\circ$. The proposed framework using OMC-free calibration achieved significantly lower cross-subject MAE than all six baselines, while OMC-assisted calibration further reduced MAE (Fig.~\ref{fig:baseline-comparison}, $p<0.05$). Removing FiLM or TCM, or individually excluding acceleration, angular velocity, or orientation, significantly increased MAE (Fig.~\ref{fig:ablation-comparison}, $p<0.05$).
Among all joint-angle components, the lowest error was observed for pelvic tilt, with an MAE of $5.24\pm2.33^\circ$, whereas the largest error occurred for radioulnar pro-supination, with an MAE of $16.82\pm6.98^\circ$ (Fig.~\ref{fig:all-joint}). Regional MAEs were $7.27\pm1.42^\circ$ for the lower limb, $7.28\pm3.00^\circ$ for the pelvis, $7.96\pm2.76^\circ$ for the upper torso, and $11.14\pm3.27^\circ$ for the upper limb. Event-aligned error distributions are shown in Fig.~\ref{fig:all-joint-phase}. Detailed joint-angle statistics are provided in Section~II of the Supplementary Material.

\begin{table}[!t]
\renewcommand{\arraystretch}{1.15}
\setlength{\tabcolsep}{2.5pt}
\caption{Correlation, Intraclass Correlation, and Bland--Altman Results for Peak Trunk--Pelvis Rotational Metrics}
\label{tab:pointtime}
\centering
\footnotesize
\begin{tabular}{lcccc}
\hline
\textbf{Variable} &
\textbf{$r$} &
\textbf{ICC [95\% CI]} &
\textbf{Bias ($^\circ$)} &
\textbf{LoA ($^\circ$)} \\
\hline
Pelvic rot.
& 0.76
& 0.75 [0.68, 0.80]
& $-2.02$
& $[-21.10,\ 17.06]$ \\

Upper torso rot.
& 0.77
& 0.76 [0.72, 0.80]
& $+0.17$
& $[-25.83,\ 26.17]$ \\

X-factor
& 0.62
& 0.55 [0.45, 0.63]
& $+4.01$
& $[-15.64,\ 23.65]$ \\
\hline
\end{tabular}

\begin{flushleft}
\footnotesize
Peak denotes the negative peak value of each variable within each swing cycle. Bias is prediction minus reference. ICC is ICC(A,1); confidence intervals account for clustering by subject.

\end{flushleft}
\end{table}

\begin{table*}[!t]
\renewcommand{\arraystretch}{1.15}
\setlength{\tabcolsep}{4pt}
\caption{Phase-specific mean absolute error of trunk--pelvis rotation and obliquity variables across swing amplitude and event-aligned phase. Error values are reported as mean $\pm$ SD across subject-level mean MAEs} (degrees)
\label{tab:phase_mae}
\centering
\scriptsize
\begin{tabular}{lccccccccc}
\hline
\multirow{2}{*}{\textbf{Variable}} &
\multicolumn{3}{c}{\textbf{Backswing}} &
\multicolumn{3}{c}{\textbf{Downswing}} &
\multicolumn{3}{c}{\textbf{Follow-through}} \\
\cmidrule(lr){2-4}
\cmidrule(lr){5-7}
\cmidrule(lr){8-10}
& \textbf{Full} & \textbf{Half} & \textbf{Quarter}
& \textbf{Full} & \textbf{Half} & \textbf{Quarter}
& \textbf{Full} & \textbf{Half} & \textbf{Quarter} \\
\hline
Pelvic rotation
& $7.66\pm6.16$ & $6.28\pm3.14$ & $4.97\pm2.32$
& $11.14\pm7.51$ & $8.59\pm3.91$ & $8.40\pm6.60$
& $12.27\pm7.26$ & $11.24\pm5.14$ & $10.99\pm7.73$ \\

Upper torso rotation
& $11.59\pm9.85$ & $8.53\pm3.42$ & $7.03\pm3.02$
& $17.45\pm10.97$ & $12.99\pm5.94$ & $12.90\pm11.11$
& $15.20\pm5.78$ & $12.87\pm5.99$ & $12.85\pm8.47$ \\

X-factor
& $7.28\pm4.76$ & $5.50\pm2.62$ & $5.65\pm3.25$
& $11.27\pm5.59$ & $9.02\pm3.28$ & $9.75\pm5.11$
& $10.38\pm4.75$ & $8.81\pm4.50$ & $8.29\pm4.39$ \\

S-factor
& $5.99\pm2.54$ & $5.28\pm2.75$ & $4.35\pm1.95$
& $8.36\pm2.61$ & $7.62\pm3.41$ & $6.89\pm2.82$
& $9.13\pm3.11$ & $8.08\pm3.25$ & $6.97\pm2.83$ \\

O-factor
& $2.88\pm1.81$ & $2.70\pm1.62$ & $2.15\pm1.26$
& $4.49\pm2.07$ & $4.03\pm2.07$ & $3.10\pm1.53$
& $4.16\pm2.12$ & $4.06\pm2.23$ & $3.70\pm2.13$ \\
\hline
\end{tabular}
\end{table*}

\begin{figure*}[!t]
\centerline{\includegraphics[width=0.88\textwidth]{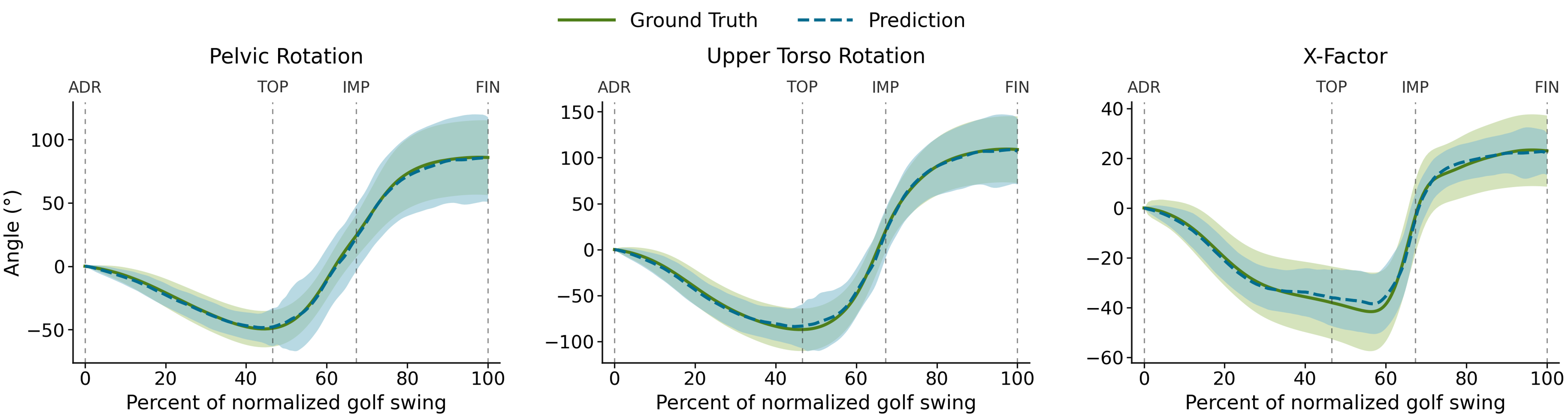}}
\caption{Event-normalized temporal profiles of pelvic rotation, upper torso rotation and X-factor during full swings. Solid lines indicate the mean OMC-derived ground truth profiles across all swings, whereas dashed lines indicate the mean smartwatch IMU-based prediction profiles. Shaded regions indicate one standard deviation across swings. All variables were zeroed at address, and the backswing, downswing, and follow-through phases were independently time-normalized before concatenation onto a common 0--100\% swing-progress axis. ADR, TOP, IMP, and FIN denote address, top of backswing, ball-impact, and finish, respectively. Both methods used the same IMU-predicted events.}
\label{fig:5-metric}
\end{figure*}

\begin{figure*}[!t]
\centerline{\includegraphics[width=0.8\textwidth]{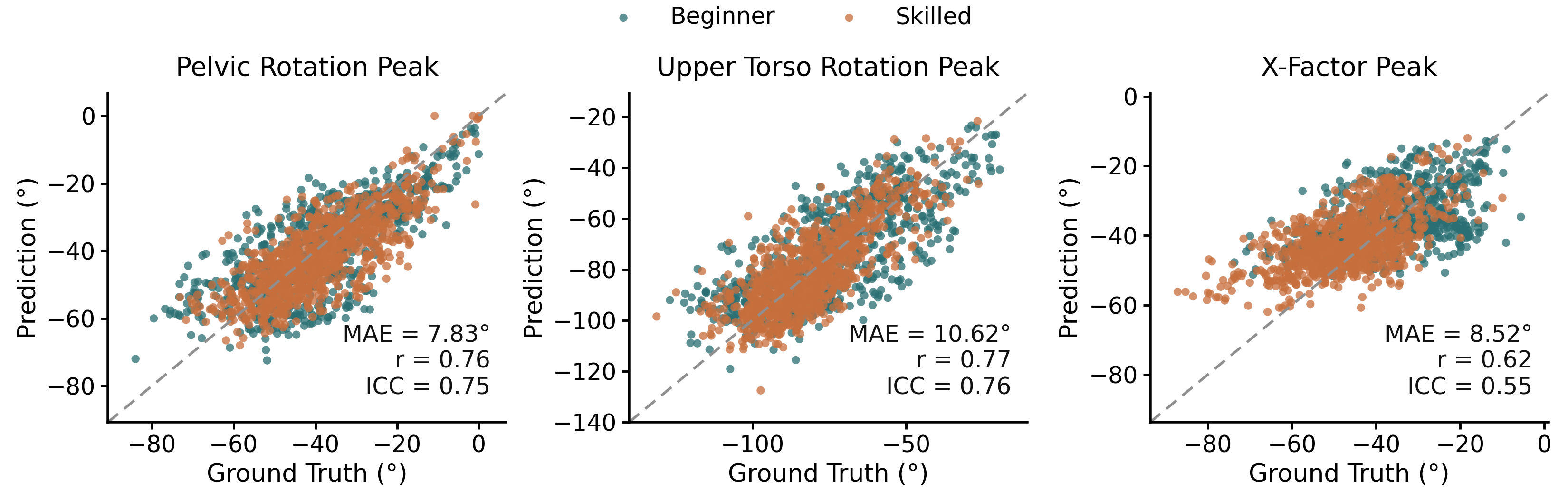}}
\caption{Correspondence between OMC-derived ground truth and smartwatch IMU-based predictions for peak values of three trunk--pelvis rotation metrics across all subjects and swings. Each point represents one swing, colors indicate skill level, and the dashed gray line denotes the line of identity.}
\label{fig:peak-points}
\end{figure*}

\begin{table*}[!t]
\renewcommand{\arraystretch}{1.15}
\setlength{\tabcolsep}{4pt}
\caption{Linear Mixed-Effects Model $p$-Values for Effects of Skill Level, Club Group, and Swing Amplitude on Swing-Level MAE of Trunk--Pelvis Kinematic Metrics}

\label{tab:lmm_error}
\centering
\footnotesize
\begin{tabular}{lccccc}
\hline
\textbf{Metric} &
\textbf{Overall MAE ($^\circ$)} &
\textbf{Skill Level ($p$)} &
\textbf{Club group ($p$)} &
\textbf{Amplitude ($p$)} &
\textbf{Significant post-hoc contrasts} \cr
\hline
Pelvic rotation
& $8.97\pm5.49$
& 0.405
& 0.866
& 0.043
& F $>$ H; F $>$ Q; H $>$ Q \\

Upper torso rotation
& $12.15\pm7.57$
& 0.777
& 1.000
& 0.014
& F $>$ H; F $>$ Q; H $>$ Q \\

X-factor
& $7.98\pm3.27$
& 1.000
& 0.405
& 0.047
& F $>$ H; F $>$ Q \\

S-factor
& $6.67\pm1.93$
& 1.000
& 1.000
& 0.001
& F $>$ H; F $>$ Q; H $>$ Q \\

O-factor
& $3.31\pm1.27$
& 0.405
& 1.000
& $<0.001$
& F $>$ Q; H $>$ Q \\
\hline
\end{tabular}
\begin{flushleft}
\footnotesize
F: full swing; H: half swing; Q: quarter swing. Club group was defined as long clubs (1W, 3W, 5H) and irons/wedges (5I, 7I, 9I, SW). Overall MAE values are reported as mean $\pm$ SD across subject-specific mean MAEs. Mixed-effects models accounted for repeated measurements within subjects. Holm correction was applied to the 15 metric--factor omnibus tests. Post-hoc contrasts were reported only for factors with significant omnibus tests and were adjusted within each metric--factor combination using the Benjamini--Hochberg false discovery rate procedure~\cite{benjamini1995controlling}. X $>$ Y indicates significantly higher MAE in group X than in group Y.
\end{flushleft}
\end{table*}

\subsection{Estimation Results of Trunk--Pelvis Metrics}
Across all swing-amplitude conditions, whole-swing subject-level MAE (mean$\pm$SD) was $8.97\pm5.49^\circ$ for pelvic rotation, $12.15\pm7.57^\circ$ for upper torso rotation, $7.98\pm3.27^\circ$ for X-factor, $6.67\pm1.93^\circ$ for S-factor, and $3.31\pm1.27^\circ$ for O-factor. Phase-specific errors were generally lower during backswing and increased during later swing phases (Table~\ref{tab:phase_mae} and Fig.~\ref{fig:5-metric}). Swing-level profile analysis yielded Fisher-aggregated Pearson correlations of 0.99 for pelvic rotation, 0.99 for upper torso rotation, 0.97 for X-factor, 0.97 for S-factor, and 0.93 for O-factor. Their respective 95\% CIs were [0.98, 0.99], [0.98, 0.99], [0.96, 0.97], [0.96, 0.98], and [0.91, 0.95]. Detailed trunk--pelvis metric error statistics are provided in Section~III of the Supplementary Material.

\subsection{Correspondence Between Estimated and Ground Truth Peak Trunk--Pelvis Rotation Metrics}
Rotational metrics at their negative peaks were evaluated using Pearson correlation, ICC, and Bland--Altman analysis (Table~\ref{tab:pointtime}). For peak metrics, upper torso rotation showed the strongest correspondence with the ground truth measurements ($r=0.77$, ICC $=0.76$), followed by pelvic rotation and X-factor ($r=0.76$ and $0.62$; ICC $=0.75$ and $0.55$) (Fig.~\ref{fig:peak-points}). Despite relatively small mean biases, LoA remained wide, particularly for upper torso rotation. At ball-impact, results of the three rotational metrics are provided in Section~VIII of the Supplementary Material.

\subsection{Effects of Skill Level, Club Group, and Swing Amplitude on Estimation Error}
Linear mixed-effects models showed that swing amplitude significantly affected the estimation error of all five golf-specific temporal variables ($p<0.05$; Table~\ref{tab:lmm_error}). Skill level and club group did not show significant effects. Full swings showed significantly higher MAE than quarter swings for all five metrics and than half swings for all except O-factor. Half swings showed significantly higher MAE than quarter swings for all except X-factor.

\section{Discussion}
\label{sec:discussion}

This paper presents the Wrist-IMU Temporal Kinematic Network (WIT-KinNet) for estimating full-body golf swing joint angles from a single wrist-worn smartwatch IMU, with OMC-free IMU calibration and IMU-based swing segmentation evaluated under laboratory conditions. The model uses IMU embedding with FiLM, followed by residual feed-forward and temporal convolution modules, to combine inertial features with temporal context. 
The results demonstrated OMC-validated full-body joint-angle reconstruction and the ability to capture temporal patterns in trunk--pelvis rotational metrics, with estimation error significantly affected by swing amplitude. These findings support the feasibility of single-smartwatch kinematic estimation using real sensor measurements. Although the MAE was higher than that reported using synthetic inertial inputs, the present study evaluated real smartwatch measurements against marker-based OMC, whereas Lauer used video-derived reference kinematics~\cite{lauer2025learning}. Compared with the trunk- and pelvis-mounted IMU configuration used by Kim et al.~\cite{kim2023validation}, the proposed framework requires only a single wrist-worn smartwatch, reducing sensor placement requirements.


\subsection{Analysis of Overall Joint-Angle Accuracy}
The proposed framework achieved an overall full-body joint-angle MAE of $8.53 \pm 2.05^\circ$ (mean$\pm$SD across subject-specific mean MAEs). The joint-dependent differences in MAE magnitude reflect the fundamental information asymmetry inherent to single-point IMU sensing (Fig.~\ref{fig:all-joint}). The lowest errors were observed in several relatively stereotyped joint-angle components, including pelvic tilt, subtalar pro-supination, lumbar bending, hip rotation, and hip adduction. In contrast, the largest errors occurred in upper-limb components, particularly radioulnar pro-supination and elbow flexion. Despite being anatomically proximate to the sensor, upper-limb rotational components exhibit high inter-individual variability~\cite{zheng2008kinematic} driven by grip style, swing plane, and lead-arm positioning, making them inherently harder to reconstruct from a single wrist-worn IMU. The regional error analysis further confirmed the joint-level error distribution. The lower limb exhibited the lowest region-level MAE, followed by the pelvis, upper torso, and upper limb. The elevated upper-limb errors are mainly attributable to rotational components such as radioulnar pro-supination, which are sensitive to highly individual technique factors that are poorly encoded by wrist inertial signals alone. Future work may benefit from incorporating anatomical constraints or kinematic priors for the upper limb during post-processing to reduce physically implausible estimates in upper-limb joint angles.
The baseline comparisons and ablations support the proposed combination of feature modulation, temporal convolution, and complementary input modalities under the evaluated conditions (Fig.~\ref{fig:baseline-comparison} and~\ref{fig:ablation-comparison}).

\subsection{Analysis of Phase-Specific Error and Biomechanical Interpretation}
Estimation errors increased during the later stages of the swing in both the full-body joint-angle analysis and the trunk--pelvis kinematic metrics (Fig.~\ref{fig:all-joint-phase}, Table~\ref{tab:phase_mae}, and Fig.~\ref{fig:5-metric}). 
The phase-dependent pattern reflects the changing biomechanical demands of the swing. The backswing corresponds to a controlled loading phase with moderate angular velocities and smoother motion trajectories, whereas the downswing spans the transition from the top of the backswing into the main acceleration phase. Prior studies have reported that the time from the top of the backswing to ball-impact is approximately 0.23--0.29~s~\cite{sanders2020relationship}, reflecting the high-rate nature of the downswing and its associated large segment angular velocities and impact-related inertial transients.
Follow-through is further affected by post-impact deceleration and unloading dynamics, which are less temporally stereotyped across golfers and swing amplitudes.
The high-energy phases are difficult to encode fully from a single wrist-worn IMU because inertial-based estimates are sensitive to movement speed, sensor range, attachment quality, and sampling frequency, especially in dynamic or impact-related tasks~\cite{chen2017effects,cereatti2024isb}. Thus, the increased errors during downswing and follow-through may reflect the combined effects of higher inertial signal magnitude, reduced temporal margin, and limited observability of whole-body motion from the wrist sensor. The same pattern was observed across full, half, and quarter swings, while the overall error magnitude decreased with swing amplitude. These phase-dependent errors could potentially be reduced by incorporating event- or phase-aware temporal modeling.

\subsection{Analysis of Temporal Correlation for Trunk--Pelvis Performance Metrics}
The high level of correlation observed for pelvic ($r=0.99$) and upper torso rotation ($r=0.99$) indicates that the proposed framework captures the broad shape of the event-normalized rotational profiles (Fig.~\ref{fig:5-metric}). 
X-factor and S-factor also demonstrated strong correlation ($r=0.97$ and $0.97$, respectively), while O-factor showed comparatively lower but still strong correlation ($r=0.93$). The comparatively lower O-factor correlation is qualitatively consistent with the dedicated-segment IMU study of Kim et~al.~\cite{kim2023validation}, although sensor placement and evaluation methods are different. 
The observed temporal correlations suggest that the estimated trunk--pelvis profiles may support post-swing visualization and review. Trunk--pelvis kinematic metrics have been associated with clubhead speed, ball velocity, and golfer skill level~\cite{hume2005role,myers2008role,chu2010relationship,meister2011rotational,zhou2022swing}. However, it is worth noting that correlation is insensitive to systematic offsets and amplitude differences, and event normalization can emphasize similarity between stereotyped swing profiles. For example, upper torso rotation showed a temporal correlation of $r=0.99$ despite a relatively large whole-swing MAE of $12.15\pm7.57^\circ$, illustrating that similar temporal profiles do not necessarily imply small angular errors.

\subsection{Analysis of Peak Trunk--Pelvis Rotation Estimates}
Peak pelvic and upper torso rotation showed stronger agreement with ground truth than X-factor (Table~\ref{tab:pointtime}), suggesting that segmental rotation amplitudes were estimated more consistently than their relative separation. Because X-factor combines upper torso and pelvic rotation, errors in either estimate can affect its value. The relatively small mean biases for pelvic and upper torso rotation indicate limited average offset, while the wider limits of agreement reveal variability in individual-swing estimates. These findings complement the temporal analysis: the framework reproduced broad rotational profiles, with greater uncertainty in the amplitude of individual peaks.

\subsection{Effects of Skill Level, Club Type, and Swing Amplitude}
The linear mixed-effects analysis showed that swing amplitude was the most pervasive modulator of estimation error, reaching statistical significance for all five performance variables ($p < 0.05$; Table~\ref{tab:lmm_error}). The swing-amplitude effect is consistent with the phase-specific analysis, as greater swing amplitude increases peak angular velocity and the contribution of high-error downswing and follow-through phases to swing-level MAE.
Skill level and club group showed no significant main effects after correction. Detailed mixed-effects results, interaction analyses, and residual diagnostics are provided in Section~VII of the Supplementary Material, in which exploratory analysis identified skill-by-amplitude interactions for pelvic rotation and O-factor, suggesting that amplitude-related changes in estimation error differed between beginners and skilled golfers.

\subsection{Practical Implications for Golf Coaching}
The proposed framework enables full-body swing kinematics to be estimated from a single smartwatch, substantially reducing the sensing requirements for biomechanical assessment. By reconstructing the full-body joint angles, the system may support phase-specific visualization and post-swing review, providing quantitative information that can complement video analysis and coach observation when discussing movement patterns with golfers.

The level of measurement precision required in coaching depends on how the reconstructed kinematics are used. Previous biomechanical studies indicate that differences in golf-related rotational measures can occur on the scale of only several degrees. For example, Meister et al.~\cite{meister2011rotational} reported a peak X-factor of $56\pm4^\circ$ in professional golfers, while Lephart et al.~\cite{lephart2007eight} reported standard errors of measurement of $2.18^\circ$, $2.79^\circ$, and $3.33^\circ$ for upper
torso rotation, pelvic rotation, and X-factor at the top of the backswing, respectively, reflecting the repeatability of their golf-swing kinematic measurements. These values provide useful context for interpreting the scale of measurement variability in golf-swing kinematics.

In the present study, the limits of agreement for peak metrics and the relatively large whole-swing MAE for upper torso rotation indicate that uncertainty in individual kinematic estimates can be substantial relative to rotational differences of only a few degrees. Therefore, the current framework is better suited to visualizing broad rotational patterns than to interpreting small changes in individual angles or peak values. In practical coaching, angle and peak estimates should consequently be considered together with the full rotational trajectories, video observations, and the golfer's overall movement pattern.

For longitudinal training applications, observed changes should be interpreted with consideration of both measurement variability and practical relevance. Distribution-based criteria, such as the $0.2$ between-subject SD convention used by Parker et al.~\cite{parker2017effects}, may provide useful context for study design. A task-specific threshold for meaningful within-golfer change has not yet been established for the present framework; determining this threshold requires repeated-session measurements and validation against training-relevant outcomes. At present, the framework is therefore best viewed as a tool for supporting quantitative interpretation of broader movement patterns rather than for defining small, individual-level performance changes.

\subsection{Limitations and Future Directions}

Several limitations should be noted. First, OMC-free calibration yielded higher MAE than OMC-assisted calibration, suggesting that further improvement in calibration accuracy is needed.
Second, the dataset included only right-handed golfers assessed in an indoor laboratory. Repeated donning, cross-day consistency, device transfer, outdoor use, and real-world magnetic disturbances were not evaluated. Automatic identification of individual swings within long-duration continuous recordings was also not implemented; the event detector localized events within predefined trial recordings. These factors may affect performance during practical deployment. Future work should extend the pipeline to continuous swing detection and evaluate the robustness of the pipeline across sessions, repeated donning, devices, and outdoor environments. Longitudinal training outcomes were also not assessed, and thresholds for detecting meaningful within-golfer changes remain to be established.
Third, the subject-wise evaluation was limited to club types and swing amplitudes represented during training. Generalization to unseen clubs, club groups, or swing amplitudes remains to be assessed using leave-one-club(-group)-out and leave-one-amplitude-out evaluations.
Fourth, the present study did not explicitly investigate how individual characteristics, such as body proportions, swing mechanics, or wrist-to-body kinematic coupling, contribute to inter-subject variability in reconstruction performance. Future work should examine these factors to better characterize model generalization to unseen individuals.


\section{Conclusion}
\label{sec:conclusion}
In this paper, we presented the Wrist-IMU Temporal Kinematic Network (WIT-KinNet) for estimating full-body golf swing kinematics from a wrist-worn IMU. The framework integrates IMU embedding with FiLM and temporal convolutional encoding and was evaluated on a synchronized dataset of smartwatch inertial signals and OMC-derived ground truth kinematics from thirty-six golfers across two skill levels, seven club types, and three swing-amplitude conditions.
The proposed framework achieved a cross-joint mean MAE of $8.53\pm2.05^\circ$ across subject-specific mean errors and yielded temporal Pearson correlations of $0.99$ for both pelvic and upper torso rotation. Peak rotational metrics showed ICCs ranging from 0.55 to 0.76, while phase-specific analysis revealed larger errors during the downswing and post-impact phases, consistent with the high-velocity dynamics near ball contact. Swing amplitude significantly affected estimation error, with larger amplitudes generally presenting greater challenges.
The overall performance and factor-specific analyses support the feasibility of single-smartwatch full-body golf swing reconstruction. The framework provides a potential solution for accessible post-swing biomechanical assessment, with future validation extending to repeated sessions and on-course use.

\bibliographystyle{IEEEtran}
\bibliography{refs}


 




\vfill

\end{document}